\documentclass[12pt]{article}

\usepackage{times}
\usepackage{helvet} 
\usepackage{courier} 
\usepackage[hyphens]{url}
\usepackage{caption}
\DeclareCaptionStyle{ruled}{labelfont=normalfont,labelsep=colon,strut=off} 
\usepackage{algorithm}

\usepackage{amsmath}
\usepackage{amsthm}
\usepackage{amssymb}
\usepackage{bbm}
\usepackage{mathtools}
\usepackage{algpseudocode}
\usepackage{lineno}
\usepackage[top=30truemm,bottom=30truemm,left=25truemm,right=25truemm]{geometry}
\usepackage{listings}
\usepackage{subcaption}
\usepackage{url}
\usepackage{subcaption}
\usepackage{booktabs}
\usepackage{array}
\newcolumntype{P}[1]{>{\centering\arraybackslash}p{#1}}

\floatstyle{ruled}
\newfloat{listing}{tb}{lst}{}
\floatname{listing}{Listing}

\title{\textbf{StyleDiff: Attribute Comparison Between Unlabeled Datasets in Latent Disentangled Space}}

\author{Keisuke Kawano, Takuro Kutsuna, Ryoko Tokuhisa, Akihiro Nakamura, Yasushi Esaki \\
{\normalsize Toyota Central R\&D Labs., Inc.}    \\ 
{\normalsize kawano@mosk.tytlabs.co.jp}
}
\date{}

\newcommand*{\coloneqq}{=}

\begin{document}

\maketitle

\begin{abstract}
One major challenge in machine learning applications is coping with mismatches between the datasets used in the development and those obtained in real-world applications.
These mismatches may lead to inaccurate predictions and errors, resulting in poor product quality and unreliable systems.
To address the mismatches, it is important to understand in what sense the two datasets differ.
In this study, we propose StyleDiff to inform developers about the types of mismatches between the two datasets in an unsupervised manner.
Given two unlabeled image datasets, StyleDiff automatically extracts latent attributes that are distributed differently between the given datasets and visualizes the differences in a human-understandable manner.
For example, for an object detection dataset, latent attributes might include the time of day, weather, and traffic congestion of an image that are not explicitly labeled.
StyleDiff helps developers understand the differences between the datasets with respect to such latent attribute distributions. 
Developers can then, for example, collect additional development data with these attributes and conduct additional tests for these attributes to enhance reliability.
We demonstrate that StyleDiff accurately detects differences between datasets and presents them in an understandable format using, for example, driving scene datasets.
\end{abstract}

\section{Introduction}
\label{sec:intro}
One of the major challenges in machine learning applications is handling the mismatches between the dataset used in the development and those obtained in real-world applications.
These mismatches cause machine learning systems that are well trained in development and highly accurate in testing to fail unexpectedly in real-world applications.
The first step to address the mismatches is to understand where the differences between the two datasets exist.
Understanding the differences allows the developers of machine learning systems to appropriately address the problem by, for example, acquiring additional data and developing a fail safe system.
For example, in developing an object detection system for autonomous vehicles, if snow day data is found to be insufficient in the development dataset, the developer can go to collect data on snow days, develop systems specifically for system for snow days, or prohibit the use on snow days.
In this study, we aim to show developers of machine learning systems in what sense the two datasets are different, especially for image datasets.
We believe that by understanding the difference between the datasets and taking the appropriate steps, such as additional tests, the developers can improve machine learning systems with high accuracy and reliability.
In this paper, we refer to the dataset obtained from the real-world application as \textit{real dataset} and that used during development as \textit{development dataset}. 
Note that the development dataset is used to train and test the models, design the model architectures, and create specifications.

To inform the differences between datasets in a human-understandable manner, we focus on \textit{attributes} in images, which are easy to understand.
For example, for an object detection system in autonomous vehicles that detects cars, attributes might include the time of day, weather, and congestion level.
By quantifying the difference between datasets for each attribute that is easy to understand, the developers can intuitively understand in what sense the datasets are mismatched.
However, such attributes are diverse and usually not explicitly labeled in the datasets.
We aim to find attributes that have different distributions in the two datasets without manually defining these attributes and labeling the attributes to each image.

Several approaches have been proposed to present the difference between two datasets.
One approach is to calculate the distance between datasets such as Wasserstein distance~\cite{alvarez2020geometric} and FID~\cite{heusel2017gans, chong2020effectively}.
Although the distances quantify the difference between datasets, they do not reveal in what sense the datasets are different.
Another approach is to select the representative images from the datasets~\cite{kawano2022partial,bachem2017practical,sener2018active}.
This approach allows us to specify images that represent the mismatches between the real and development datasets.
However, the approach still has two challenges.
\begin{itemize}
    \item The mismatches between the two datasets are only visualized as selected images, making it difficult to infer which attributes are differently distributed in these datasets.
    Furthermore, even if there are differences between datasets with multiple attributes, these methods can only provide a single image set with a mixture of images representing the multiple attributes. 
    This makes it more difficult to understand which attributes differ in distributions.
    \item Because existing methods are formulated as discrete optimization problems, some are computationally difficult to apply to large datasets containing tens of thousands of images. For example, the computational complexity of PWC~\cite{kawano2022partial} and the coreset approach (k-center greedy)\footnote{Note that for the dataset covering problems~\cite{kawano2022partial}, the distances between $N$ candidate points in one dataset and $N$ centers in the other dataset are required.}~\cite{bachem2017practical,sener2018active} are $O(N^2)$, where $N$ is the number of images in the datasets.
\end{itemize}

\begin{figure}[t]
    \centering
    \includegraphics[height=11cm]{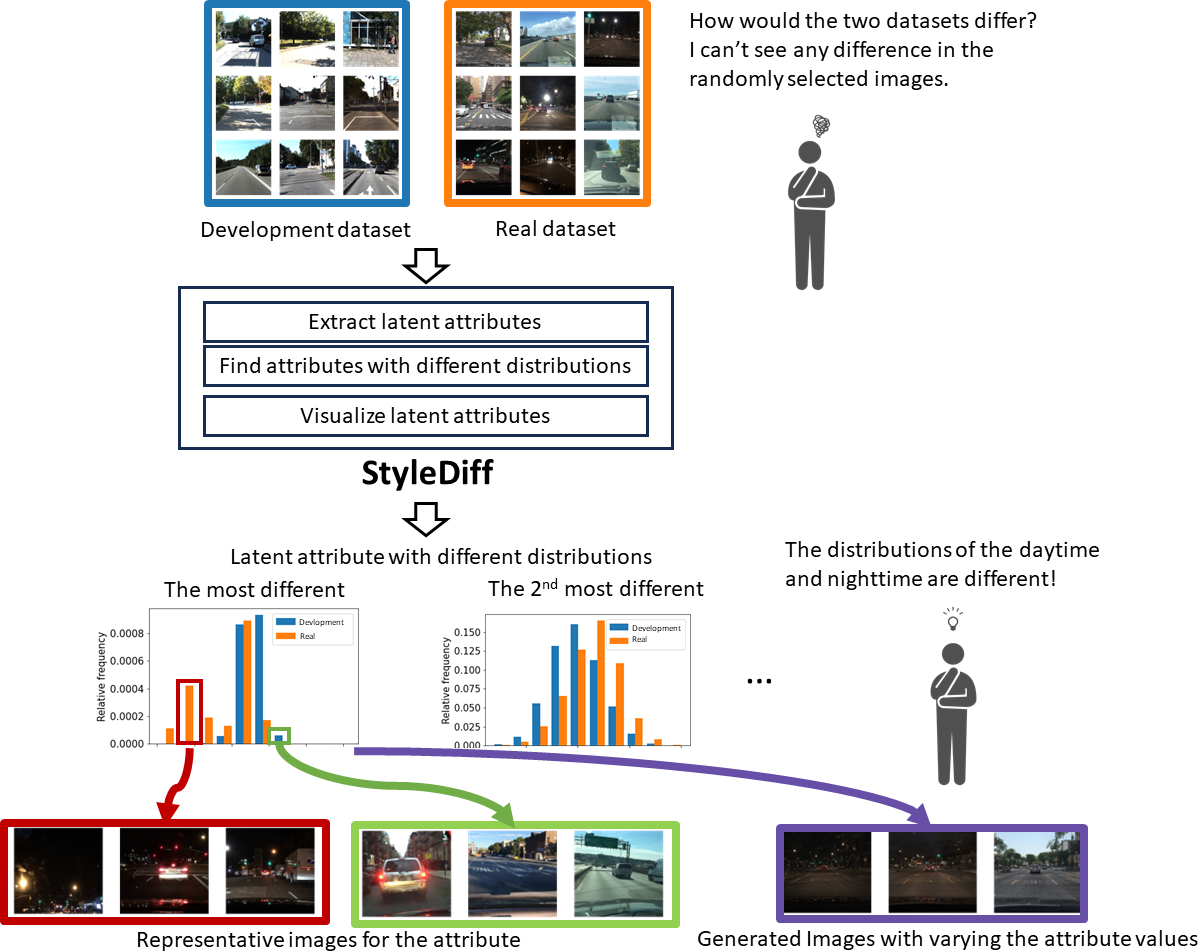}
    \caption{Concept of StyleDiff. StyleDiff compares the distributions of latent attributes extracted from the two input datasets. StyleDiff presents information about the latent attributes that differ in their distributions using the histograms, representative images, and images generated by changing the attribute.}
    \label{fig:fig1}
\end{figure}

In this study, we propose StyleDiff to present how the two datasets differ by comparing attribute distributions in the datasets.
We demonstrate the concept of StyleDiff in Figure~\ref{fig:fig1}.
StyleDiff first extracts latent attributes from the images in the real and development datasets.
Then, StyleDiff quantifies the differences between the distributions of extracted attributes.
Finally, StyleDiff visualizes the attributes that differ in their distributions by using histograms of the distributions, selecting representative images, and generating images with different attribute values.
As shown in Figure~\ref{fig:fig1}, the histogram illustrates the distributions of the extracted attribute.
The histogram allows the developers to know how the attribute is distributed in the two datasets.
The developers can also display the representative images, which locates at specific regions in the histogram, as shown in red and green in Figure~\ref{fig:fig1}.
The purple in Figure~\ref{fig:fig1} shows an image sequence generated by StyleDiff, where only the attribute varies.
The representative images and the image sequence help the developers to understand what the extracted attribute is.
Even if there are differences between multiple attribute distributions, StyleDiff visualizes these attributes individually by selecting the attributes and providing images for each of the selected attributes.

One difficulty with implementing StyleDiff is that the two datasets do not contain explicit information about the attributes.
Therefore, to realize StyleDiff, it is necessary to extract the attributes from the datasets.
To extract the various attribute from each image, we propose using the disentangled space obtained from the generative models such as StyleGANs~\cite{karras2019style,karras2020stylegan2, wu2021stylespace}.
Each dimension of the disentangled space is expected to correspond to an attribute in the images~\cite{wu2021stylespace}.
By quantifying the difference between two datasets for each dimension of the disentangled space, StyleDiff compares the distributions for each attribute that is not explicitly labeled in the datasets.
Then, StyleDiff presents the attributes corresponding to the distributions with the large mismatches to the developers.
Presenting mismatches for each attribute helps developers intuitively understand the mismatches between datasets.

Our contributions are summarized as follows.
\begin{itemize}
    \item We propose StyleDiff to present how the two datasets differ by comparing attribute distributions. StyleDiff automatically extracts latent attributes with different distributions and presents the differences in a human-understandable manner (Section~\ref{sec:StyleDiff}).
    \item We show that the computational complexity of StyleDiff is $O(d N \log N)$, where $N$ is the number of images and $d$ is the number of extracted attributes if a pretrained encoder is available, allowing StyleDiff to be applied to large datasets (Section~\ref{sec:computational_complexity}).
    \item Experimentally, we demonstrate that StyleDiff outperforms existing methods in extracting images that correspond to the differences between the datasets (Section~\ref{sec:quantitative}).
    Furthermore, using real-world datasets, we demonstrate that StyleDiff can present the attributes individually, even if there are differences among multiple attributes (Section~\ref{sec:exp3}).
\end{itemize}

The rest of this paper is organized as follows.
In Section~\ref{sec:StyleDiff}, we describe our proposed StyleDiff.
Section~\ref{sec:related} presents a literal review of related work.
Subsequently, in Section~\ref{sec:experiment}, we show some numerical experiments to demonstrate StyleDiff.
Finally, in Section~\ref{sec:conclusion}, we present our conclusion.

\section{StyleDiff}
\label{sec:StyleDiff}

\begin{figure}[t]
    \centering
    \includegraphics[width=0.99\linewidth]{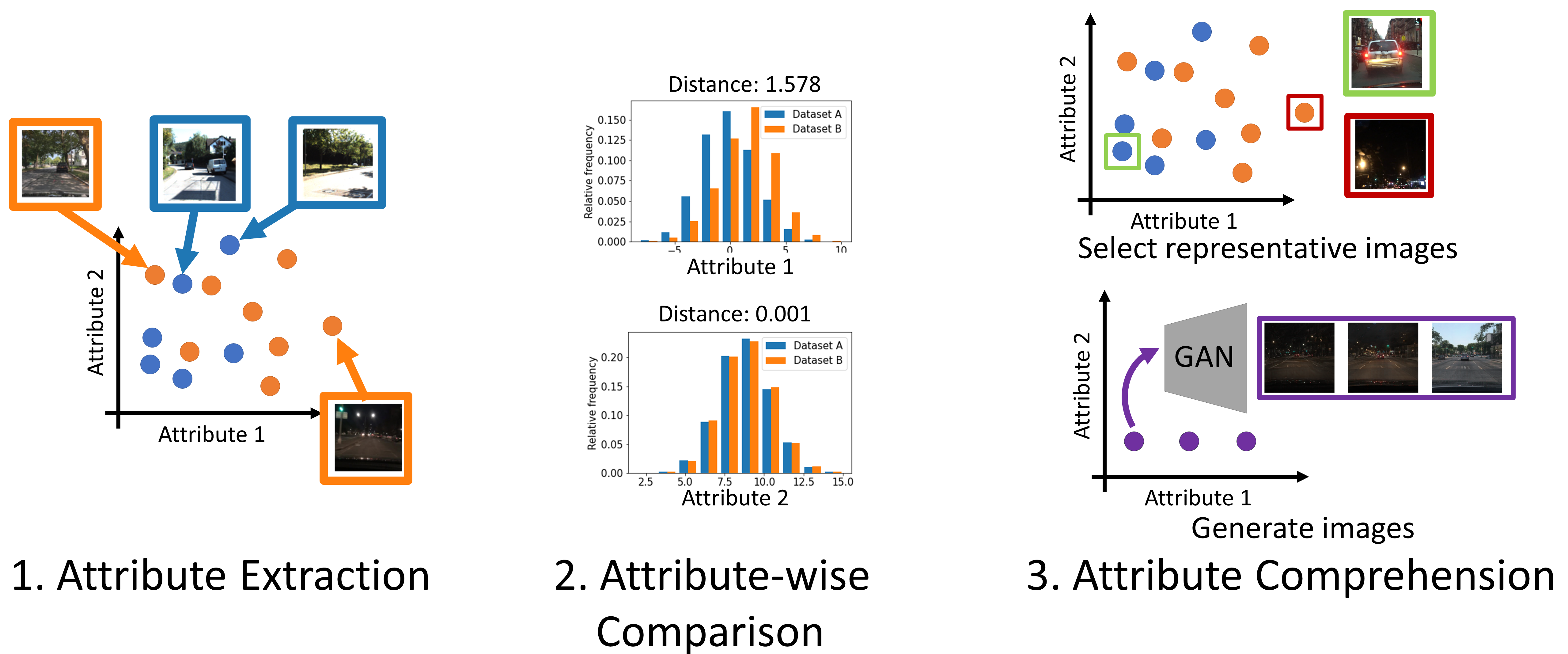}
    \caption{Overview of StyleDiff. StyleDiff first extracts attribute distributions from the two unlabeled datasets. The distributions are then compared to identify the attributes that have large differences.
        Finally, to show the users the identified attributes, StyleDiff presents attributes by selecting some images at the specific regions in the distributions and/or generating images by changing the attribute.}
    \label{fig:method}
\end{figure}

Given two unlabeled image datasets, StyleDiff automatically extracts latent attributes with different distributions between the datasets and visualizes the differences in a human-understandable manner.
StyleDiff leverages the embedding space of images, where each dimension ideally has a one-to-one correspondence with a single attribute; that is each dimension of the embedding space controls a single attribute~(disentanglement) and each attribute is controlled by a single dimension~(completeness).
We call the embedding space \textit{attribute space} and the embedding vectors \textit{attribute vectors}.
For the attribute space, we employ the latent space of the recently developed GAN~\cite{karras2020stylegan2, alaluf2021restyle}, which is known for its high disentanglement and completeness~\cite{wu2021stylespace}.
Using the attribute space, StyleDiff extracts the attributes that differ in their distributions and presents them in an easy-to-understand form.
As illustrated in Figure~\ref{fig:method}, StyleDiff comprises the following three steps.
\begin{enumerate}
    \item Attribute Extraction: transform each image in the two datasets into an attribute vector.
    \item Attribute-Wise Comparison: calculate the distances between the distributions for each attribute using the attribute vectors and select attributes based on the distances.
    \item Attribute Comprehension: visualize the selected attributes by selecting images from the datasets according to the distributions of the attributes or by generating images using a generator.
\end{enumerate}
In the following, we describe each of them individually.

\paragraph{Notations}
Vectors and matrices are denoted in bold style (e.g., $\mathbf x, \mathbf A$).
$\mathbf 1_N \in \mathbb R^N$ represents an $N$-dimensional vector with all elements being ones.
$\mathbb R_+$ is a set of real positive numbers.
The $L_p$ norm is denoted as $\|\cdot \|_p$.
For a natural number $N$, $[\![N ]\!]$ means $\{1, \dots, N\}$.
$\delta_{a}$ is the Dirac delta at $a\in \mathbb R$.

\subsection{Attribute Extraction}
\label{sec:acquire}
The proposed method converts all images $\mathbf x^{(i)}, \mathbf y^{(j)} \in \mathbb R^{H\times W\times 3}$ in two datasets $A\coloneqq \{\mathbf x^{(i)} \mid i\in [\![N_x ]\!] \}$ and $B \coloneqq \{ \mathbf y^{(j)} \mid j\in [\![N_y ]\!] \}$ into attribute vectors $\mathbf s_{x}^{(i)}, \mathbf s_{y}^{(j)} \in \mathbb R^d$, where $d$ denotes the number of dimensions for the attribute vectors.
As attribute vectors, we can use style vectors~\cite{wu2021stylespace} obtained from StyleGAN2~\cite{karras2020stylegan2} and Restyle encoder~\cite{alaluf2021restyle} trained on the same domain as the two datasets.

Style vectors, intermediate latent vectors of StyleGAN2, are renowned for their disentangled and complete latent space called StyleSpace~\cite{karras2020stylegan2}.
Let $\mathbf z\in \mathbb R^{d_\text{in}}$ be a latent vector of StyleGAN2, where $d_\text{in}$ is the dimension of the vector, and StyleGAN2 first transforms it into the style vector using a feedforward neural network $f:\mathbb R^{d_\text{in}} \mapsto \mathbb R^d$ as $f(\mathbf z)$.
Subsequently, StyleGAN2 generates an image from the style vector as $G(f(\mathbf z))$, where $G: \mathbb R^d \mapsto \mathbb R^{H\times W\times 3}$ is the generator function.
To obtain the style vectors corresponding to the images in the datasets, we employ the Restyle encoder~\cite{alaluf2021restyle} $h:\mathbb R^{H\times W\times 3} \mapsto \mathbb R^{d_\text{in}}$, which predicts the latent vectors $\mathbf z$ corresponding to the input image $\mathbf x$.

We denote the function to convert images into attribute vectors as $\phi: \mathbb R^{H\times W\times 3}\mapsto \mathbb R^{d}$.
By applying $\phi$ to images $\mathbf x^{(i)}$ and $\mathbf y^{(j)}$, the following two sets of attribute vectors can be obtained.
\begin{linenomath*}
\begin{align}
    \label{eq:acquire-style-vectors}
    \mathcal S_x  \coloneqq \{\mathbf s_{x}^{(i)} \mid i\in [\![N_x ]\!] \}, \quad \mathcal S_y  \coloneqq \{\mathbf s_{y}^{(j)} \mid j\in [\![N_y ]\!] \},
\end{align}
\end{linenomath*}
where $\mathbf s_{x}^{(i)} = \phi(\mathbf x^{(i)})$ and $\mathbf s_{y}^{(j)} = \phi(\mathbf y^{(j)})$.
With the composite function of the Restyle encoder and the feedforward network of StyleGAN2 $\phi = f\circ h: \mathbb R^{H\times W\times 3} \mapsto \mathbb R^{d}$, we can obtain the style vectors corresponding to the images in the datasets.

\subsection{Attribute-Wise Comparison}
\label{sec:attribute-wise-comparison}
StyleDiff calculates the distance between the two sets of attribute vectors for each dimension of the attribute space.
Because each dimension of the attribute space is expected to correspond to an attribute in the images, this process is equivalent to calculating the distance between the two distributions for each attribute.
For the sets of attribute vectors $\mathcal S_x$ and $\mathcal S_y$, we extract the $c$-th dimensional values and define the following two empirical distributions $X_c$ and $Y_c$ for each dimension.
\begin{linenomath*}
\begin{align}
    X_c \coloneqq \frac{1}{N_x}\sum_{i=1}^{N_x}\delta_{s_{x,c}^{(i)}} \quad
    Y_c \coloneqq \frac{1}{N_y}\sum_{j=1}^{N_y}\delta_{s_{y,c}^{(j)}},
\end{align}
\end{linenomath*}
where $s_{x,c}^{(i)}, s_{y,c}^{(j)} \in \mathbb R$ are $c$-th values of $\mathbf s_{x}^{(i)}$ and $\mathbf s_{y}^{(j)}$, respectively.

StyleDiff employs the Wasserstein distance $\mathcal W(X_c, Y_c)$~\cite{peyre2019computational} to evaluate the dissimilarities between the empirical distributions.
The Wasserstein distance requires no additional hyperparameters, such as kernels, and can be obtained rapidly, as described in Section~\ref{sec:computational_complexity}.
The Wasserstein distance between the empirical distributions is defined as follows:
\begin{linenomath*}
\begin{align}
    \label{eq:wass}
     & \mathcal {W}(X_c, Y_c) \coloneqq  \left(\min_{\mathbf P\in U(\mathbf m_x, \mathbf m_y)} \sum_{i=1}^{N_x}\sum_{j=1}^{N_y} P_{ij}  D_{c, ij} \right)^{\frac{1}{2}}, \\    &
    \ U(\mathbf m_x, \mathbf m_y) = \{ \mathbf P \in [0,1]^{N_x \times N_y} \ | \ \mathbf P^\top \mathbf 1_{N_x}= \mathbf m_y, \ \mathbf P \mathbf 1_{N_y} = \mathbf m_x \},
\end{align}
\end{linenomath*}
where $\mathbf P$ is the matrix representing the transport plan, and its $(i,j)$-element $P_{ij}$ represents the mass transported from  $\mathbf x^{(i)}$ to $\mathbf y^{(j)}$.
$\mathbf m_x \coloneqq [\frac{1}{N_x}, \dots, \frac{1}{N_x}]^\top \in \mathbb R_{+}^{N_x}$ and $\mathbf m_y \coloneqq [\frac{1}{N_y}, \dots, \frac{1}{N_y}]^\top \in \mathbb R_{+}^{N_y}$ are the masses for the images.
$U(\mathbf m_x, \mathbf m_y)$ corresponds the mass conservation constraints.
$D_{c, ij}$ is the cost of transporting the mass from $s_{x,c}^{(i)}$ to $s_{y,c}^{(j)}$.
For example, the following squared Euclidean distance is often used:
\begin{linenomath*}
\begin{align}
\label{eq:transport-cost}
D_{c, ij} = \|\mathbf s_{x,c}^{(i)}-\mathbf s_{y, c}^{(j)}\|^2_2
\end{align}
\end{linenomath*}
StyleDiff calculates the distances $\mathcal W(X_c, Y_c)$ for all $c\in [\![d]\!]$, and selects $K$ dimensions with the largest distances.
We denote the selected dimensions sorted by the distances as $\mathbf c^* = [c^*_{1}, \dots, c^*_{K}]$.

We propose normalizing distances between attribute vectors based on their scales, as the scales have a greater impact on the distance than the attribute distributions' differences if the scales differ significantly (see Section~\ref{sec:quantitative}).
We can normalize the cost of transportation (Eq.~\eqref{eq:transport-cost}) as $D_{ij} = \frac{1}{\sigma_c^2} | s_{x,c}^{(i)} - s_{y,c}^{(j)} |^2$, where $\sigma_c$ is the standard deviation of $\{s_{x,c}^{(i)} \mid i\in [\![N_x]\!]\} \cup \{s_{y,c}^{(j)} \mid j\in [\![N_y]\!]\}$.
This normalization aligns the scale of the attribute vectors on all dimensions, eliminating the influence of the scales.

\subsection{Attribute Comprehension}
The selected dimensions $\mathbf c^*$ are expected to correspond to attributes in the images.
To understand the attributes corresponding to each selected dimension $c^*_l \in \mathbf c^*$, we propose two methods as shown in Figure~\ref{fig:fig1}.
One is selecting representative images from the two datasets based on the $c^*_l$-th values of the attribute vectors~(i.e., $s_{x,c^*_l}^{(i)}$ and $s_{y,c^*_l}^{(j)}$).
The representative images are for example, images corresponding to the endpoints with maximum or minimum values, or images corresponding to the region where the difference between the two distributions is significant.
The other is generating an image sequence, in which only the corresponding attribute is varied, by changing the $c^*_l$-th value of an attribute vector using the StyleGAN2~\cite{karras2020stylegan2}.
An image sequence in which only one attribute is varied clearly informs the user which attribute corresponds to the $c^*_l$-th dimension of the attribute space.
StyleDiff can be used for the dataset covering problems~\cite{kawano2022partial} by selecting representative images, such as images at the endpoint of the dimension with the largest distance.

\subsection{Attribute Vector Modification}
We experimentally found that an attribute may correspond to multiple dimensions of StyleSpace depending on the learned domain.
If an attribute corresponds to multiple dimensions, i.e., low completeness, a problem in visualizing the attributes may occur, as described in Section~\ref{sec:exp3}.
We propose mitigating this problem by applying principal component analysis~(PCA) to the style vectors, as in~\cite{harkonen2020ganspace}.
By applying PCA, highly correlated dimensions in the style vectors are combined into one.
Note that PCA does not only increases the completeness but also decreases the disentanglement, i.e., several correlated attributes may be combined into a single dimension.

\subsection{Computational Complexity}
\label{sec:computational_complexity}
Let $N=\max\{N_x, N_y\}$ be the number of images in the datasets and $K$ be the number of dimensions to select.
Given a function $\phi$ that converts each image into an attribute vector, the computational complexity of converting images into attribute vectors is $O(N_x + N_y)$.
The dimension-wise Wasserstein distances can be calculated by $O(dN \log N)$~\cite{peyre2019computational} since $s_{x,c}^{(i)}$ and $s_{y,c}^{(j)}$ are scalar values.
In addition, visualization of the selected attributes requires $O(K)$ for image generation, and $O(N)$ for image selection, (e.g., images with minimum and maximum values).
Therefore, the total computational complexity of the proposed method is $O(d N \log N)$, enabling StyleDiff to be applied to large datasets.

StyleDiff requires the function $\phi$ such as Restyle encoder~\cite{alaluf2021restyle} trained in the same domain as the two target datasets.
It is challenging to accurately estimate the amount of computation needed for training GAN encoders; however, we emphasize that they have been effectively trained on datasets of over 10,000 images in practice.

\section{Related Work}
\label{sec:related}
\paragraph{Dataset covering}

The problem of selecting the representative images from datasets to represent the difference between the datasets is formulated as a dataset covering problem~\cite{kawano2022partial}.
Methods for addressing the dataset covering the problem include PWC, anomaly detection, and active learning methods.
PWC visualizes which patterns in one dataset are missing in another dataset by selecting data points to minimize the partial Wasserstein divergence~\cite{peyre2019computational} between the real dataset and union of the development dataset and selected data points.
Anomaly detection methods~\cite{breunig2000lof, scholkopf1999support, kim2019rapp, an2015variational} can be considered  a method of detecting the difference between datasets in terms of detecting data in one dataset that is missing in another.
Active learning methods~\cite{holub2008entropy, coletta2019combining, sener2018active, shui2020deep} can also be used to extract data in one dataset that cannot be correctly predicted by a model trained on another dataset.
Unlike these methods, StyleDiff aims to compare the two datasets by their attribute distributions, and to visualize the difference between them.
The proposed method enables us to select and generate images corresponding to attributes and to compare the histograms of the attributes.
This rich information is useful for developers to investigate missing patterns in the development dataset.

\paragraph{Geometric dataset distance}
Several metrics such as Wasserstein distance~\cite{alvarez2020geometric} and Fr\'echet Inception Distance (FID)~\cite{heusel2017gans} have been proposed as distances between two datasets.
Wasserstein distance between datasets is used, for example, to evaluate the difficulty of transfer learning~\cite{alvarez2020geometric} and to find correspondences between data in cross-lingual datasets~\cite{alvarez2018gromov}.
FID and its variants~\cite{heusel2017gans,chong2020effectively} have been proposed to evaluate the qualities of images generated by GANs~\cite{goodfellow2014gan}.
FID is defined as the Fr\'echet distance~\cite{dowson1982frechet} between two Gaussian distributions of image feature vectors.
It is known that the FID is equivalent to the Wasserstein distance between two Gaussian distributions~\cite{heusel2017gans}.

The Wasserstein distance is defined as a linear optimization problem, as in Eq.~\eqref{eq:wass}.
This optimization problem can be solved, for example, using the interior-point method~\cite{peyre2019computational} and Sinkhorn iterations~\cite{cuturi2013sinkhorn, benamou2015iterative}.
In the worst-case scenario, their computational complexity is $O(N^3)$ and $O(TN^2)$, respectively, where $N$ is the number of data points and $T$ is the number of iterations.
As a special case, it is also known that the Wasserstein distance can be computed in $O(N \log N)$ if the dimension of vectors $d$ is equal to 1~\cite{peyre2019computational}.

Unlike these distances between datasets, StyleDiff aims to present users with information on what differences exist between the two datasets.
To this end, StyleDiff quantifies the difference between the extracted attribute distributions instead of directly quantifying the difference between the datasets.
Furthermore, StyleDiff helps developers intuitively understand the mismatches by visualizing the attributes by, for example, generating images.
We emphasize that only the distance between datasets does not provide a detailed understanding of the mismatches.

\paragraph{StyleSpace}
StyleSpace, which is a disentangled space for images~\cite{wu2021stylespace, harkonen2020ganspace}, has been used for several tasks~\cite{alaluf2021restyle, patashnik2021styleclip, alaluf2022times, lang2021explaining, chong2021retrieve}.
Image manipulation is a popular application of the StyleSpace~\cite{alaluf2021restyle, patashnik2021styleclip, alaluf2022times}.
By changing some elements of the style vector corresponding to an input image and decoding it, we can obtain an image in which only some attributes are edited.
Lang et al., employed StyleSpace to illustrate a trained classifier~\cite{lang2021explaining} as another example of an application.
StyleDiff also uses the disentangled StyleSpace for the attribute-wise comparison of two datasets, however, our objective (i.e., comparing two datasets) is distinct from these methods.

\section{Experiment}
\label{sec:experiment}

In this section, we demonstrate how StyleDiff determines and visualizes the difference in attributes between the given two datasets through numerical experiments.
First, we apply StyleDiff to two artificially generated datasets with different distributions of an attribute to demonstrate the functionality of the proposed method.
A quantitative evaluation of the ability to identify attributes with varying distributions was conducted by applying the proposed and existing methods to artificially generated datasets.
Finally, we illustrate StyleDiff in a realistic scenario by comparing two driving scene datasets.
The implementations of StyleGAN2~\cite{karras2020stylegan2} and Restyle~\cite{alaluf2021restyle} used in the following experiments are the publicly available ones~\cite{stylegan2implementation,restyleimplementation}, respectively.
The training of StyleGAN2 was performed on an NVIDIA Tesla A100, and all other computations were performed on an NVIDIA RTX TITAN.
All hyperparameters used during training are default values of the public implementations~\cite{stylegan2implementation,restyleimplementation}.

\subsection{Demonstration: Artificially Generated Datasets}
\label{sec:exp1}
In this subsection, we demonstrate how the attributes are extracted and visualized by the proposed method using two datasets that are artificially generated to ensure that their distributions in an attribute are different.
The two datasets are generated by randomly sampling 5,000 images for each from the FFHQ dataset~\cite{karras2019style}, which is a dataset of human face images.
To generate the two datasets, we use the publicly available attributes~\footnote{\url{https://github.com/DCGM/ffhq-features-dataset}} of the FFHQ dataset.
There are 34 attributes estimated from the images, including categorical values (e.g., beard) and continuous values (e.g., age).

\begin{figure}
    \centering
    \begin{subfigure}{0.45\textwidth}
        \includegraphics[width=\textwidth]{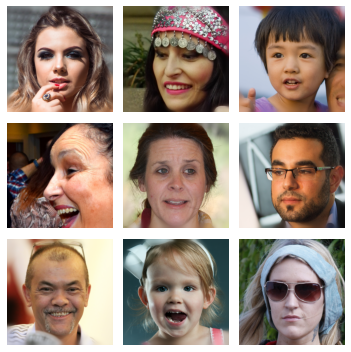}
        \caption{Dataset $A$}
    \end{subfigure}\ \ \ \  \
    \begin{subfigure}{0.45\textwidth}
        \includegraphics[width=\textwidth]{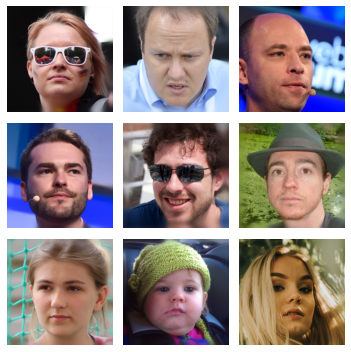}
        \caption{Dataset $B$}
    \end{subfigure}
    \caption{Examples randomly drawn from two generated datasets (FFHQ~\cite{karras2019style})}
    \label{fig:sample}
\end{figure}

We generate two datasets with different distributions of smiling, which is one of the attributes.
One dataset $A$ contains 75\% of smiling people and 25\% of non-smiling people, while another dataset $B$ contains 25\% of smiling people and 75\% of non-smiling people.
In this experiment, we convert the degrees of smiling (i.e., continuous values from 0 to 1) into a binary label with a threshold of 0.5.
Figure~\ref{fig:sample} illustrates examples that are randomly drawn from the two generated datasets.
As shown in Figure~\ref{fig:sample}, the images contain various attributes (e.g., face orientation, hairstyle, and brightness), rendering it challenging to distinguish the two datasets merely by examining the examples.

Given the two generated datasets, we demonstrate how the datasets are compared and how the attribute (i.e., smiling) is detected and visualized by StyleDiff.
In the experiment, the attribute vectors are style vectors obtained from the publicly available pretrained Restyle encoder~\cite{alaluf2021restyle, stylegan2implementation} and StyleGAN2~\cite{karras2020stylegan2, restyleimplementation}, which are trained on the FFHQ dataset.
The dimension of the attribute space ($d$) is 6,048.

\begin{figure}
    \centering
    \includegraphics[width=0.45\linewidth]{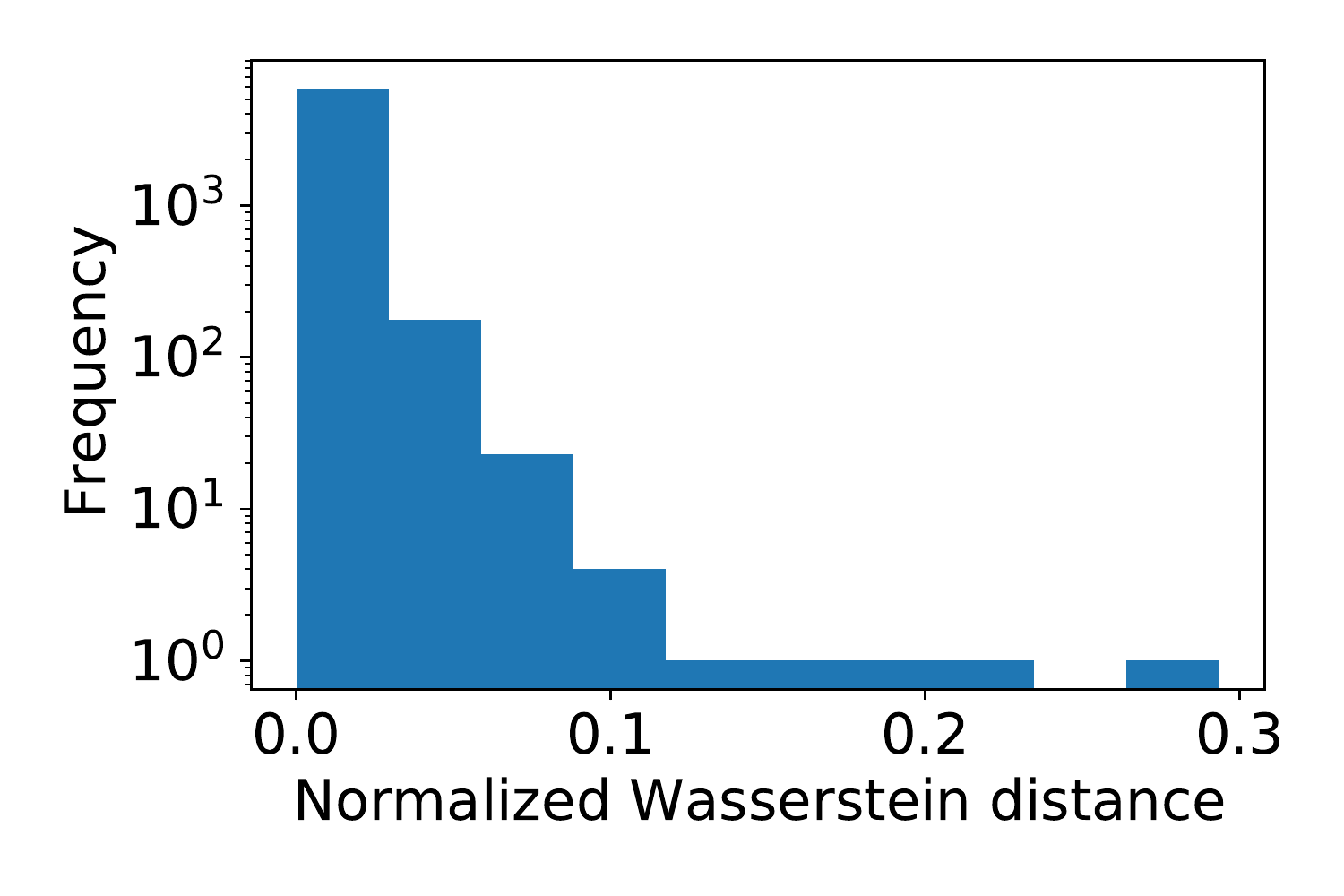}
    \caption{Histogram of dimension-wise distances between two datasets}
    \label{fig:frequency}
\end{figure}

\begin{figure}
    \centering
    \begin{subfigure}{0.47\textwidth}
        \includegraphics[width=\textwidth]{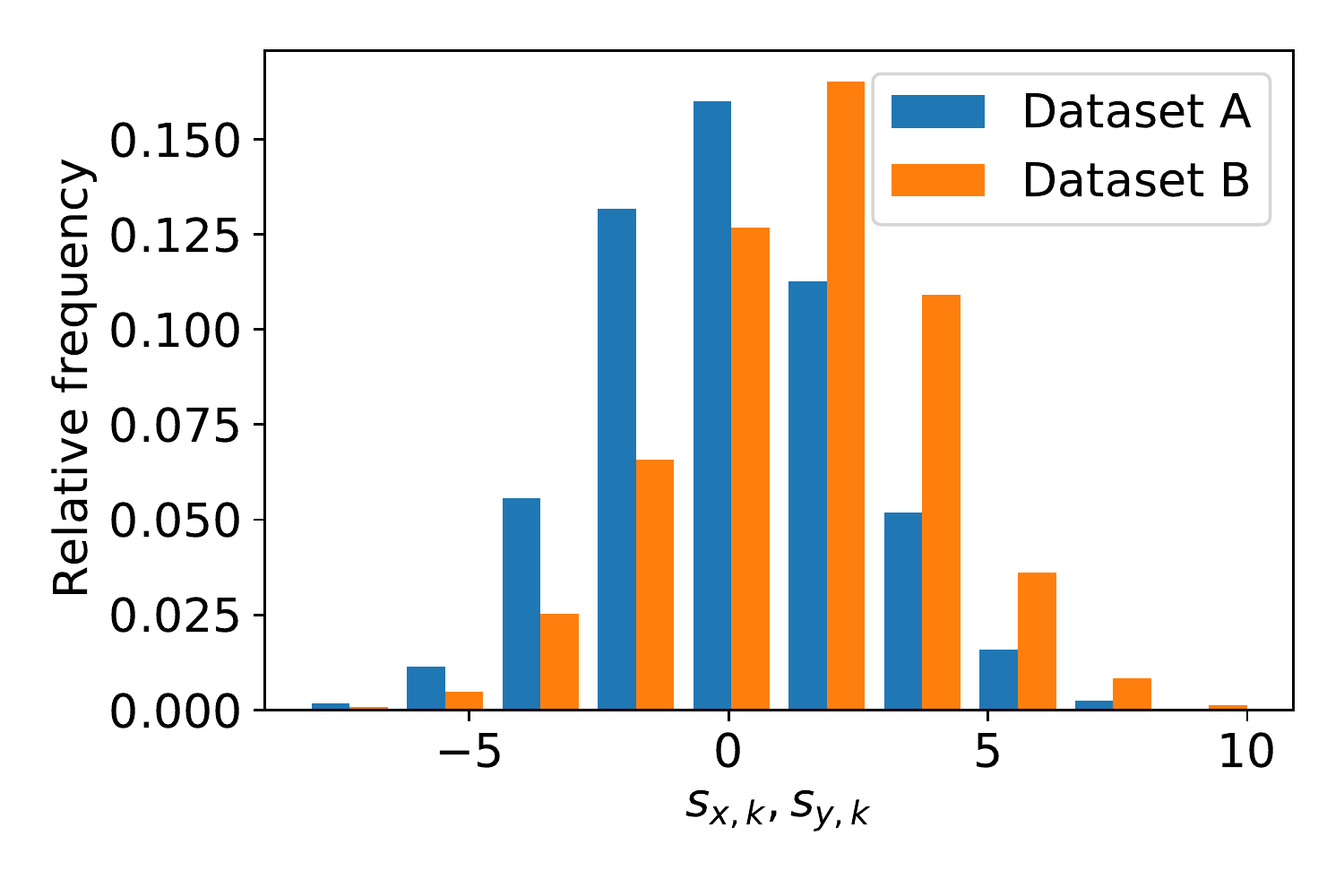}
        \caption{Distribution of the attribute vectors on the dimension with the largest distance}
    \end{subfigure}\ \ \ \  \
    \begin{subfigure}{0.47\textwidth}
        \includegraphics[width=\textwidth]{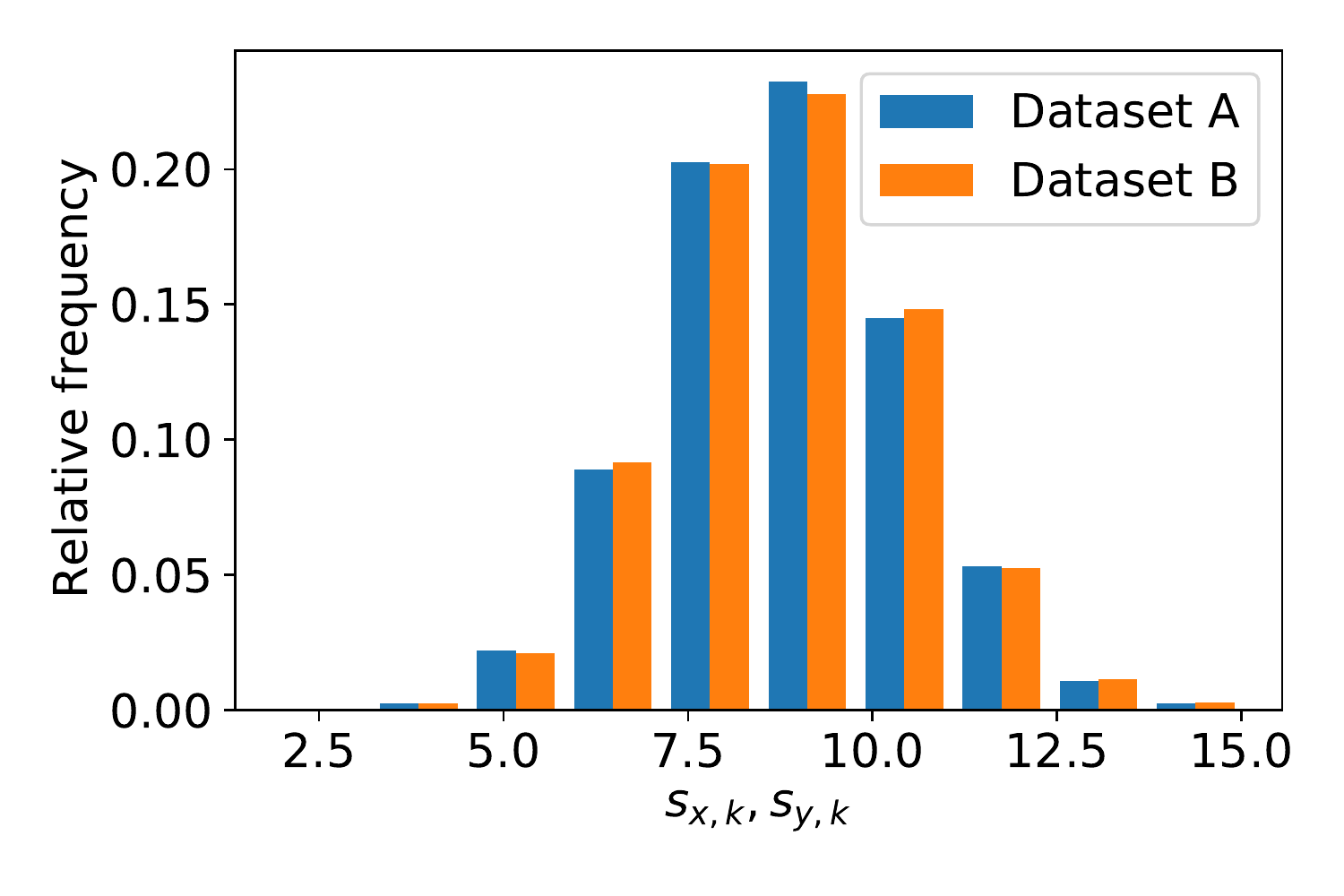}
        \caption{Distribution of the attribute vectors on the dimension with the smallest distance}
    \end{subfigure}
    \caption{Distribution of the attribute vectors}
    \label{fig:attributed_vectors}
\end{figure}

After transforming each image in the two datasets into an attribute vector, StyleDiff determines the distances between the set of the attribute vectors $\mathcal S_x, \mathcal S_y$ for each dimension of the attribute space.
The histogram of the dimension-wise distances between two datasets is shown in Figure~\ref{fig:frequency}.
As illustrated in Figure~\ref{fig:frequency}, only two dimensions have values above 0.2 while the remaining 6046 dimensions have values close to zero.
This implies that the attribute vectors have high completeness; that is, an attribute is controlled by a few dimensions of the attribute space. 
We also illustrate the distribution of the attribute vectors on the dimension with the largest distance (Wasserstein distance: 0.293) and on the dimension with the smallest distance (Wasserstein distance: 0.000) in Figure~\ref{fig:attributed_vectors}.
As shown in Figure~\ref{fig:attributed_vectors}, the difference between the two distributions can be quantified using the Wasserstein distance, which can be rapidly computed for a set of scalar values.

\begin{figure}
    \centering
    \begin{subfigure}{0.45\textwidth}
        \includegraphics[width=\textwidth]{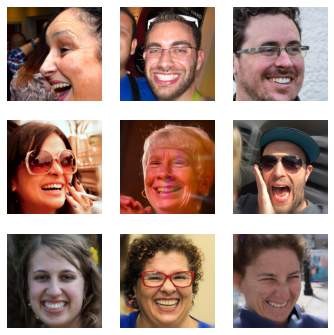}
        \caption{Images with low values on the selected dimension}
    \end{subfigure}\ \ \ \  \
    \begin{subfigure}{0.45\textwidth}
        \includegraphics[width=\textwidth]{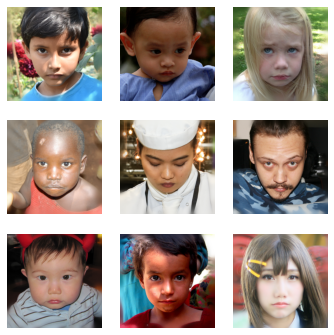}
        \caption{Images with high values on the selected dimension}
    \end{subfigure}
    \caption{Results of StyleDiff: Attribute comprehension via representative images. The images correspond to the endpoints of the dimension with the largest distance (Fig.~\ref{fig:attributed_vectors}~(a)).}
    \label{fig:endpoint}
\end{figure}

StyleDiff allows us to select dimensions by distance and visualize the corresponding attributes.
Here, for example, we select the dimension with the largest distance and visualize the corresponding attribute.
The images that exist at the endpoints of the selected dimension are shown in Figure~\ref{fig:endpoint}.
As shown in Figures~\ref{fig:endpoint} (a) and (b), the images in Figure~\ref{fig:endpoint}~(a) are all smiling images, while those in Figure~\ref{fig:endpoint}~(b) are all non-smiling images.
The rest of the attributes vary in Figures~\ref{fig:endpoint}~(a) and (b).
This indicates that the selected dimension corresponds to the smile attribute.
We can assess the bias of the attribute (Smiling) in the two datasets by comparing their distributions in Figure~\ref{fig:attributed_vectors}~(a).

\begin{figure}
    \centering
    \begin{subfigure}{0.95\textwidth}
    \centering
        \includegraphics[width=0.2\textwidth]{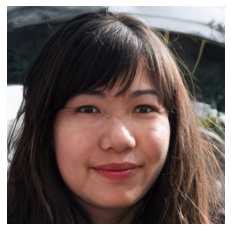}
    \caption{Generated image with the random latent vector}
    \end{subfigure}\\
    \begin{subfigure}{0.95\textwidth}
    \centering
    \includegraphics[width=0.95\linewidth]{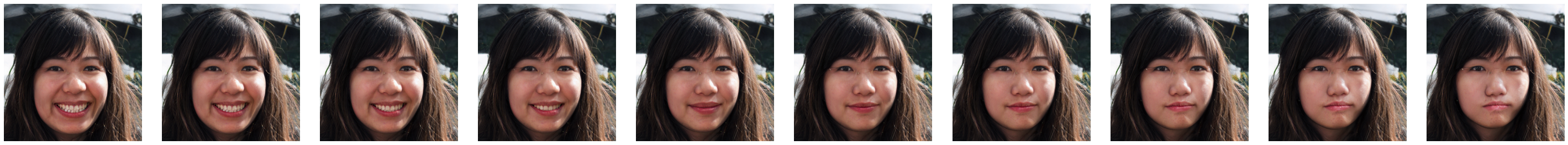}
    \caption{Generated image sequence by gradually changing the values of the selected dimension of the corresponding attribute vector}
    \end{subfigure}
    \caption{Results of StyleDiff: Attribute comprehension via generated images}
    \label{fig:random_generated_concat}
\end{figure}

We then demonstrate the visualization of the attribute by generating an image sequence.
Figure~\ref{fig:random_generated_concat}~(a) is a generated image obtained by the trained StyleGAN2~\cite{karras2020stylegan2} with a random input vector.
By gradually changing the values in the selected dimension of the corresponding attribute vector, Figure~\ref{fig:random_generated_concat}~(b) can be obtained.
In Figure~\ref{fig:random_generated_concat}~(b), only one attribute, i.e., the degree of smiling, varies gradually, while all the other attributes (e.g., background, face orientation, and hairstyle) are invariant.
The image sequence reveals that the selected dimension corresponds to the degree of smiling.
As shown in this demonstration, StyleDiff allows us to identify attributes with different distributions between two datasets that could not be discovered by only looking at the examples in Figure~\ref{fig:sample}.

\subsection{Quantitative Evaluation}
\label{sec:quantitative}

We quantitatively compare StyleDiff and existing methods~(PWC~\cite{kawano2022partial}, LOF~\cite{breunig2000lof}, coreset (k-center greedy)~\cite{sener2018active}, and FID greedy~\cite{heusel2017gans, chong2020effectively}) in terms of their ability to detect differences between two unlabeled datasets.
For the evaluation, we quantify the ability to select the images from the development dataset corresponding to the attribute that is less included in the development dataset than in the real dataset by following~\cite{kawano2022partial}.
To apply StyleDiff to the dataset covering the problem, we select the dimension with the largest distance in the distributions and then select the images in the real dataset at one of the endpoints.
To determine which of the two endpoints to select, we compare the means of the two distributions on the dimension and select the endpoint that is closer to the mean of the real dataset distribution than that of the development dataset distribution.

For the evaluation, we similarly generate two datasets to the method in Section~\ref{sec:exp1}.
A subset of 500 images sampled from the FFHQ dataset~\cite{karras2020stylegan2} was used for each of the datasets.
The development dataset and the real dataset are sampled so that the proportions of the binary labels $l \in \{0,1\}$ (e.g., smiling and non-smiling) are $p:(1-p)$ and $(1-p):p$, where $p\in [0, 0.5]$, respectively.
For the binary labels, continuously-valued attributes are scaled from 0 to 1 and then converted to binary labels with a threshold of 0.5.
For categorical values (e.g., NoGlasses, ReadingGlasses, and Sunglasses for the glasses attribute), we select the most common one for the binary label $l=0$, and the others for $l=1$.
If both continuous and categorical values are available for an attribute, the continuous value is used for the label.
If there are fewer than 500 images for a binary label, the corresponding attributes are not evaluated.
We evaluate a total of 25 attributes.

We apply the proposed StyleDiff and existing methods to the two datasets and then quantitatively evaluate their ability to detect the attributes that are less included in the development dataset.
As a metric, we use $r_p^{(l)}$, which is the percentage of images with the label $l$ in the $10$ images selected from the real dataset.
We compare the averages of $r_p^{(l)}$, i.e.,$\frac{1}{2}\sum_{l\in\{0,1\}} \frac{1}{11}\sum_{p\in \{0, 0.05, \dots, 0.5\} }r_p^{(l)}$. 

For StyleDiff, all images in the two datasets are converted into the style vectors using the trained StyleGAN2~\cite{karras2020stylegan2, stylegan2implementation} and the trained Restyle encoder~\cite{alaluf2021restyle, restyleimplementation}, and used as the attribute vectors as in Section~\ref{sec:exp1}.
In addition to the proposed method, we evaluate existing methods, PWC (PW-sensitivity-ent)~\cite{kawano2022partial}, LOF~\cite{breunig2000lof}, and coreset(k-center greedy)~\cite{sener2018active}, using the same datasets.
According to~\cite{kawano2022partial}, we use pretrained ResNet50~\cite{marcel2010torchvision} for the embedding vectors required by existing methods.
In addition to these existing methods, we also employ a greedy method using FID~\cite{heusel2017gans, chong2020effectively} (FID greedy).
Similar to PWC~\cite{kawano2022partial}, FID greedy adds data points greedily so as to minimize the FID between the real dataset and the union of the selected points and the development dataset. 
As an ablation, we also evaluate the proposed StyleDiff, where the distances are not scaled according to the scales of the attribute vectors (SD w/o normalize).
The expectation of the evaluation score is 0.75.

\begin{figure}
    \centering
    \includegraphics[width=0.55\linewidth]{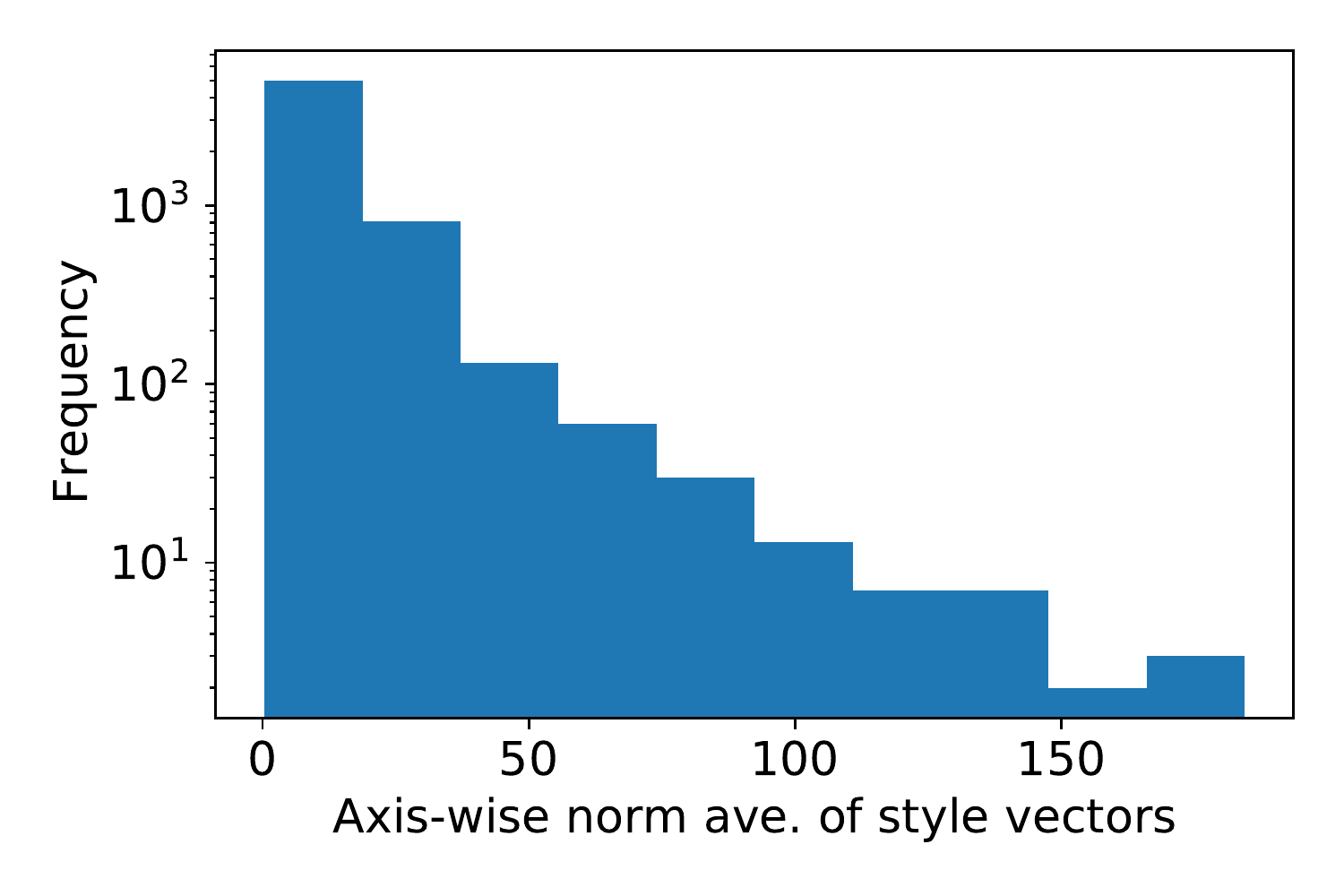}
    \caption{Average norms of the attribute vectors for each dimension}
    \label{fig:norm_dist}
\end{figure}

\begin{table}
    \caption{Quantitative evaluation results. SD denotes StyleDiff.}
    \label{tab:result}
    \centering
    \begin{tabular}{P{3.2cm}P{1.1cm}P{1.9cm}P{1cm}P{1cm}P{1.7cm}P{1cm}}
        Attribute Name        & SD             & SD w/o normalize & PWC \cite{kawano2022partial} & LOF \cite{breunig2000lof} & k-center \cite{bachem2017practical,sener2018active} &FID greedy \cite{heusel2017gans, chong2020effectively} \\	     \toprule
Smile              & \textbf{0.882} & 0.813 & 0.774 & 0.753 & 0.755 & 0.770  \\
Age                & \textbf{0.911} & 0.871 & 0.805 & 0.752 & 0.744 & 0.786  \\      
Blur               & \textbf{0.770} & 0.752 & 0.768 & 0.748 & 0.750 & 0.768  \\       
Exposure           & \textbf{0.855} & 0.836 & 0.771 & 0.753 & 0.740 & 0.773  \\           
Gender             & 0.917 & \textbf{0.921} & 0.810 & 0.741 & 0.756 & 0.784  \\         
EyeMakeup          & \textbf{0.885} & 0.883 & 0.807 & 0.753 & 0.759 & 0.789  \\            
LipMakeup          & \textbf{0.875} & 0.843 & 0.805 & 0.759 & 0.767 & 0.789  \\            
Glasses            & 0.920 & \textbf{0.929} & 0.797 & 0.748 & 0.745 & 0.791  \\          
Moustache          & 0.915 & \textbf{0.920} & 0.815 & 0.760 & 0.770 & 0.795  \\            
Beard              & \textbf{0.911} & 0.910 & 0.802 & 0.730 & 0.742 & 0.793  \\        
Sideburns          & \textbf{0.930} & 0.917 & 0.795 & 0.730 & 0.737 & 0.788  \\            
Noise              & \textbf{0.807} & 0.778 & 0.766 & 0.750 & 0.759 & 0.766  \\        
ForeheadOccluded   & \textbf{0.921} & 0.898 & 0.837 & 0.744 & 0.754 & 0.799  \\                   
HairInvisible      & \textbf{0.937} & 0.894 & 0.836 & 0.744 & 0.765 & 0.792  \\                
HairColorBrown     & \textbf{0.759} & 0.750 & 0.742 & 0.755 & 0.741 & 0.757  \\                 
HairColorBlond     & 0.752 & 0.753 & 0.738 & 0.734 & 0.718 & \textbf{0.756}  \\                 
HairColorBlack     & \textbf{0.786} & 0.748 & 0.769 & 0.753 & 0.767 & 0.773  \\                 
HairColorRed       & \textbf{0.810} & 0.779 & 0.747 & 0.744 & 0.737 & 0.788  \\               
HairColorGray      & 0.810 & \textbf{0.830} & 0.785 & 0.735 & 0.742 & 0.789  \\                
HeadPosePitch      & \textbf{0.880} & 0.835 & 0.750 & 0.751 & 0.739 & 0.761  \\                
HeadPoseRoll       & \textbf{0.813} & 0.810 & 0.751 & 0.749 & 0.759 & 0.753  \\               
HeadPoseYaw        & \textbf{0.927} & 0.924 & 0.749 & 0.755 & 0.746 & 0.756  \\              
EmotionHappiness   & \textbf{0.882} & 0.813 & 0.773 & 0.753 & 0.755 & 0.767  \\                   
EmotionNeutral     & \textbf{0.884} & 0.814 & 0.756 & 0.748 & 0.742 & 0.775  \\                 
EmotionSurprise    & \textbf{0.872} & 0.865 & 0.786 & 0.755 & 0.762 & 0.780                  
    \end{tabular}
\end{table}
The average of the scores for 10 trials with different random seeds for the sampling is presented in Table~\ref{tab:result}.
Evidently, StyleDiff outperformed the existing methods in 24 of the 25 attributes.
This indicates that StyleDiff can extract the less included attributes with higher accuracy than the existing methods, including PWC~\cite{kawano2022partial}.
The ablation study shows that SD with normalization (SD) outperformed SD without normalization (SD w/o normalization) in 20 of the 25 attributes, indicating that scaling improves the performance in finding the target attributes.
Figure~\ref{fig:norm_dist} presents the histogram of the average norms for each dimension of the 10,000 randomly sampled attribute vectors.
Figure~\ref{fig:norm_dist} indicates that the norms vary widely among the dimensions, with a few dimensions above 100 and most around 10.
Normalizing the distances removes the effect of the norms and improves the performance of the proposed method.
The results of the same experiment using CelebA dataset~\cite{liu2015faceattributes} are shown in \ref{appendix:celeba}.

\subsection{Comparing Two Driving Scene Datasets}
\label{sec:exp3}

\begin{figure}
    \centering
    \begin{subfigure}{0.45\textwidth}
        \includegraphics[width=\textwidth]{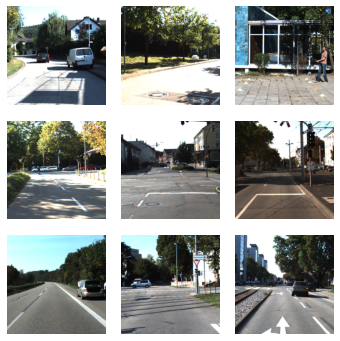}
        \caption{KITTI dataset~\cite{Geiger2013IJRR}}
    \end{subfigure}\ \ \ \  \
    \begin{subfigure}{0.45\textwidth}
        \includegraphics[width=\textwidth]{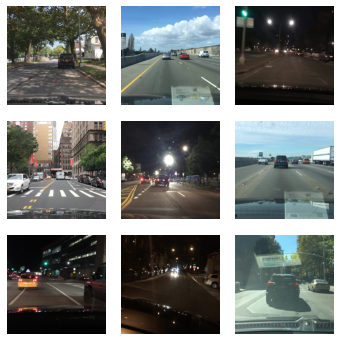}
        \caption{BDD100k dataset~\cite{bdd100k}}
    \end{subfigure}
    \caption{Images randomly drawn from KITTI and BDD100k datasets}
    \label{fig:examples}
\end{figure}

Finally, following to~\cite{kawano2022partial}, we demonstrate how StyleDiff extracts and visualizes the differences between the two datasets in a realistic situation by applying StyleDiff to widely-used driving scene datasets, KITTI~(Object Detection Evaluation 2012)~\cite{Geiger2013IJRR} and BDD100k~\cite{bdd100k}.
Both of the datasets contain images captured by vehicle-mounted cameras and are used to train and evaluate object detection models.
Figure~\ref{fig:examples} illustrates images randomly sampled from each dataset.
An obvious difference between the two datasets is the time of day the images were taken.
KITTI contains only daytime images, while BDD100k contains both daytime and nighttime images.
Since the two datasets were created by different organizations and taken in different locations, it is expected that they differ in other senses too.
However, it is not clear what other attributes differ between the two datasets.
In this experiment, we test whether the attribute that differs in the distributions (i.e., time of day) is correctly extracted by the proposed StyleDiff.
We also show the other attributes extracted by the proposed method.
The two datasets are compared using all images in the test sets of BDD100k and KITTI.
The number of images is 20,000 and 7518, respectively.

To convert the images into the attribute vectors, we trained StyleGAN2~\cite{karras2020stylegan2} and Restyle encoder~\cite{alaluf2021restyle} using all the images in the training splits of the BDD100k~\cite{bdd100k} and KITTI~\cite{Geiger2013IJRR} datasets.
As described below, the style vectors obtained from the trained models were not suitable for the attribute vectors because multiple dimensions in the vectors correspond to the same attributes.
In this experiment, we apply PCA to the style vectors and use the transformed vectors as the attribute vectors.
The number of principal components is set so that the total contribution is 0.99999.
In addition to the proposed method, we apply the existing methods, PWC (PW-sensitivity-ent)~\cite{kawano2022partial}, LOF~\cite{breunig2000lof}, and coreset(k-center greedy)~\cite{sener2018active} to the two datasets.
All hyperparameters follow to~\cite{kawano2022partial}.
For PWC~\cite{kawano2022partial}, we sampled 500 data from each of the two datasets and used them as the input due to GPU memory constraints.
We emphasize that StyleDiff can be applied to datasets with over 10,000 images without any sampling processes.

In the following, we first visualize the top-3 attributes with large distances obtained by StyleDiff.
As shown below, each of these attributes corresponds to the time of day, background, and lights at night.
Then we present the results of existing methods. 
Finally, as a limitation of the proposed method, we demonstrate the results of StyleDiff without PCA for the attribute vectors.

\begin{figure}[tp]
    \centering
    \includegraphics[width=0.99\linewidth]{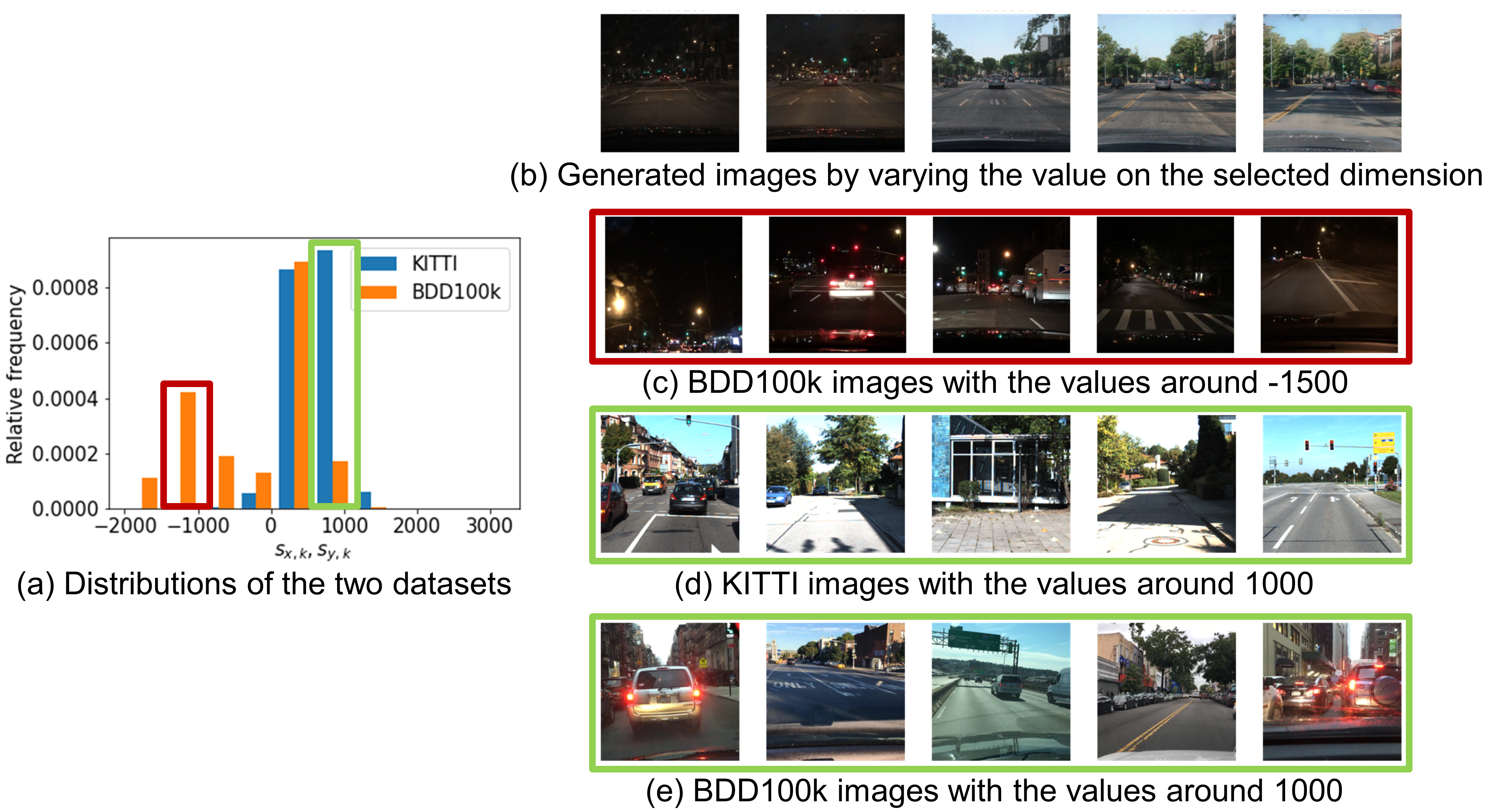}
    \caption{Results of StyleDiff: The dimension with the largest distance $(c^*_{1})$}
    \label{fig:dim1_concat}
\end{figure}

\paragraph{Visualizing the dimension with the largest distance}
The distributions of the two datasets on the dimension with the largest distance are presented in Figure~\ref{fig:dim1_concat}~(a).
We also show the image sequence generated by varying the value of the selected dimension in Figure~\ref{fig:dim1_concat}~(b), where the maximum and minimum values for the generation are the mean value $\pm$ the standard deviation for the union of the two distributions.
Figure~\ref{fig:dim1_concat}~(b), where the images vary between daytime and nighttime as the values on the dimension vary, indicates that the dimension corresponds to the difference between daytime and nighttime.
Figure~\ref{fig:dim1_concat}~(a) also shows that the region around -1500 of this dimension contains only BDD100k images.
To visualize the region, we randomly select five BDD100k images with values around -1500 and present them in Figure~\ref{fig:dim1_concat}~(c).
We also show the images of BDD100k and KITTI with values around 1000 in Figures~\ref{fig:dim1_concat}~(d) and~(e), where KITTI images are more common than BDD100k images.
Figures~\ref{fig:dim1_concat}~(c), (d), and (e) indicate that the region from -1800 to -500 corresponds to nighttime while the region around 1000 corresponds to daytime.
Clearly, StyleDiff can extract the major differences between the BDD100k and KITTI datasets (i.e., daytime and nighttime) by comparing their attribute vectors.

\begin{figure}[tp]
    \centering
    \includegraphics[width=0.99\linewidth]{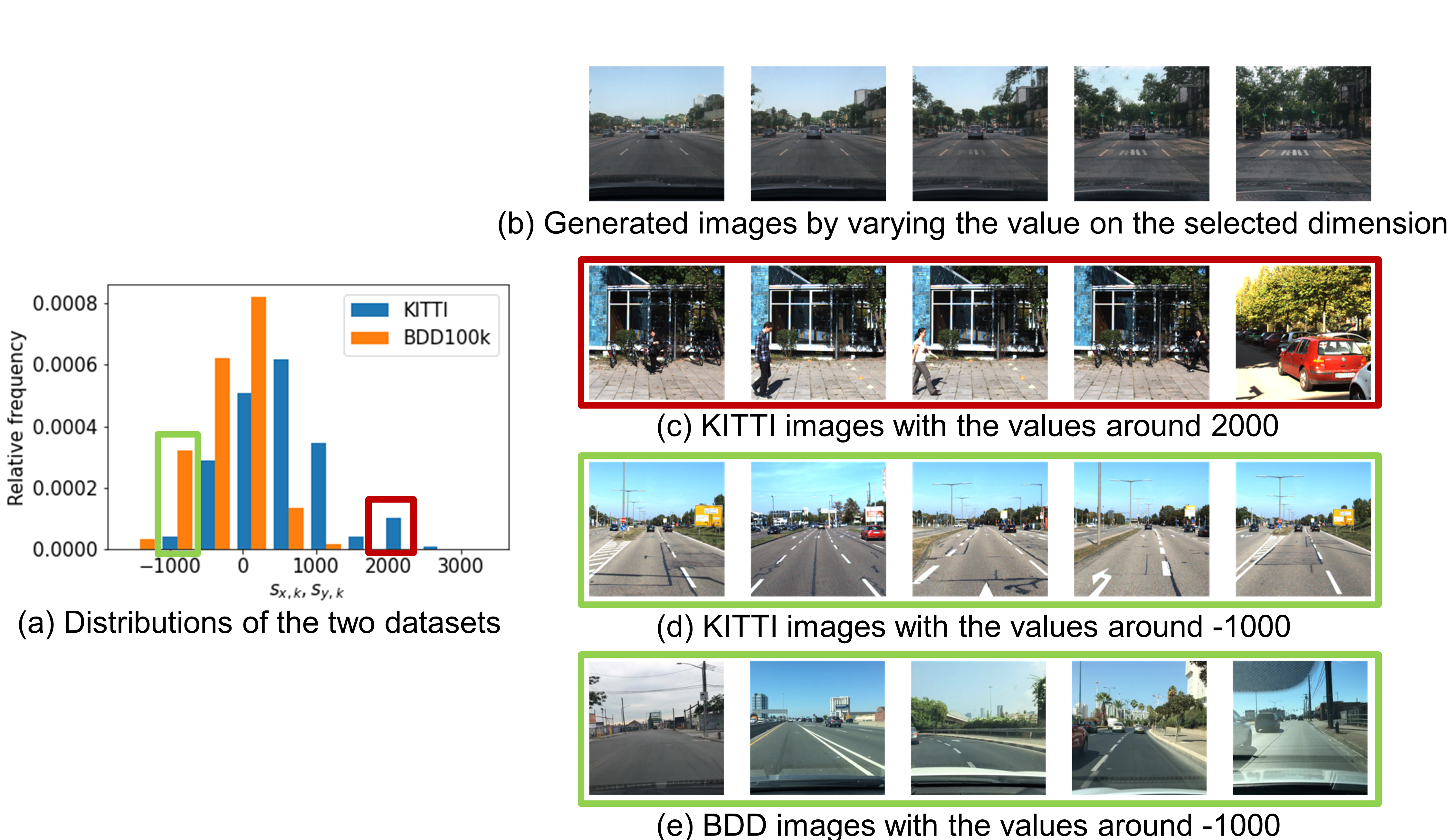}
    \caption{Results of StyleDiff: The dimension with the second-largest distance $(c^*_{2})$}
    \label{fig:dim2_concat}
\end{figure}

\paragraph{Visualizing the dimension with the second-largest distance}
We then visualize the attribute corresponding to the dimension with the second-largest distance.
The histogram of the two distributions on this dimension and a sequence of images are presented in Figures~\ref{fig:dim2_concat}~(a) and (b), respectively.
The generated images in Figure~\ref{fig:dim2_concat}~(b) indicate that the dimension corresponds to the density of the background area.
The histogram also shows that the region around 2000 contains only KITTI images.
Most of the images corresponding to this region are shown in Figure~\ref{fig:dim2_concat}~(c), illustrating the same building.
For the comparison, we also show some images corresponding to the region around -1000 in Figures~\ref{fig:dim2_concat}~(d) and~(e).
Figures~\ref{fig:dim2_concat}~(d) and~(e) illustrate the open scenes with no buildings in front.
From these visualizations, we found that BDD100k and KITTI differ in their distributions of the background because the KITTI dataset contains images captured from the vehicle stopped in front of the building.
The proposed StyleDiff can extract attributes that differ from the first attribute (daytime and nighttime), although it is impossible to separately visualize multiple attributes using existing methods that only provide a set of images.

\paragraph{Visualizing the dimension with the third-largest distance}

\begin{figure}[tp]
    \centering
    \includegraphics[width=0.99\linewidth]{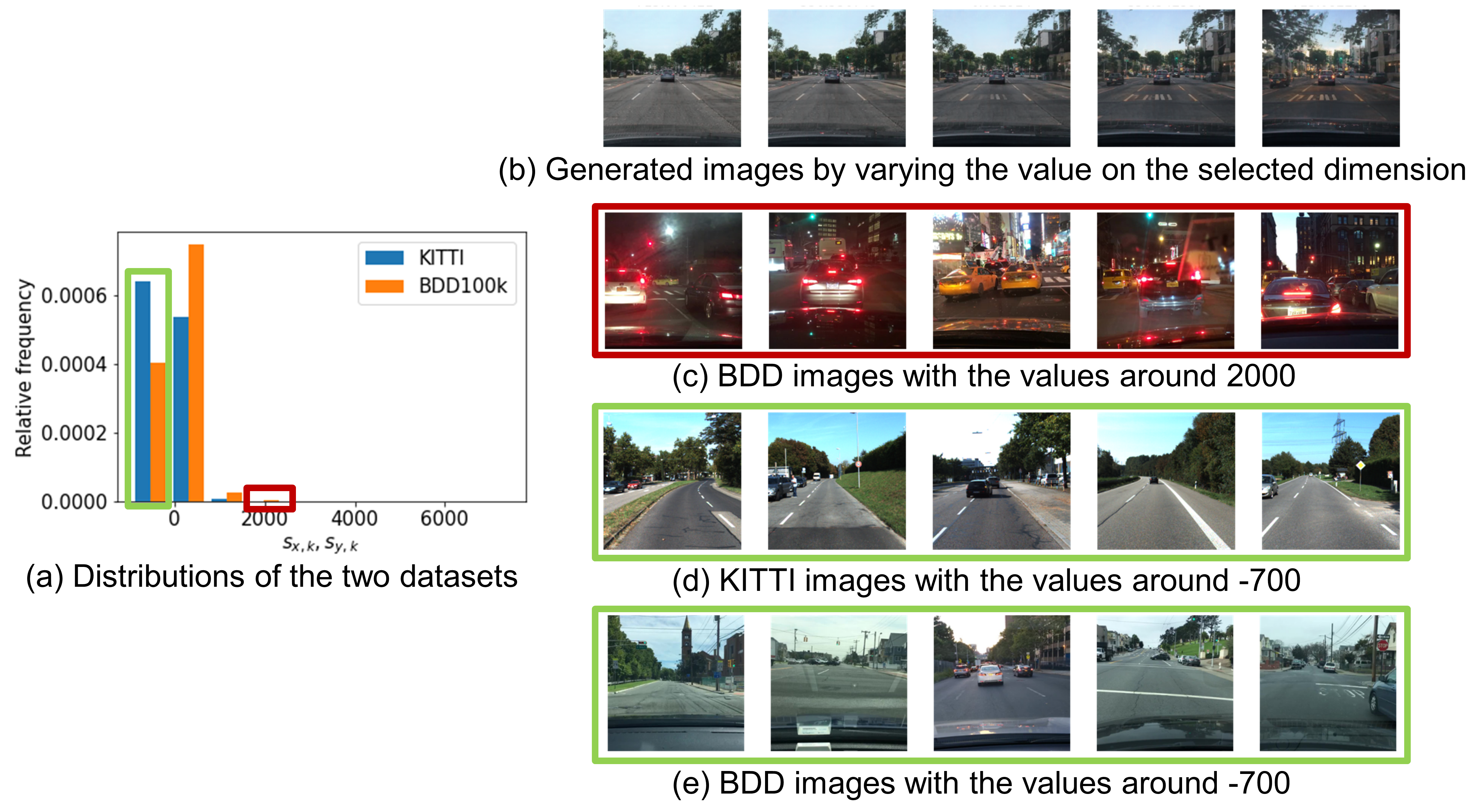}
    \caption{Results of StyleDiff: The dimension with the third-largest distance $(c^*_{3})$}
    \label{fig:dim3_concat}
\end{figure}

Figure~\ref{fig:dim3_concat}~(a) illustrates the distributions of the two datasets on the dimension with the third-largest distance.
We also demonstrate the generated image sequence corresponding to the dimension in Figure~\ref{fig:dim3_concat}~(b).
From the generated images in Figure~\ref{fig:dim3_concat}~(b), the lights in the image, such as traffic lights and tail lights, become more intense as the value of the dimension increases.
Figure~\ref{fig:dim3_concat}~(a) also indicates that the region above 2000 on the dimension contains one KITTI image and 73 BDD100k images.
The images of BDD100k with values around 2000 on the dimension are presented in Figure~\ref{fig:dim3_concat}~(c).
For comparison, the images of KITTI and BDD100k with values around -700 are shown in Figures~\ref{fig:dim3_concat}~(d) and~(e).
These figures indicate that the third dimension also corresponds to the attribute representing daytime and nighttime, similar to the first dimension.
However, unlike the first dimension, which corresponds to the night scenes without lights, the third dimension corresponds to night scenes illuminated by artificial lights (e.g., traffic and tail lights).
StyleDiff allows us to visualize similar attributes separately if they are represented on different dimensions in the attribute vectors.

As shown in Figures 10 and 12, StyleDiff successfully extracts the known attributes, i.e., the time of day.
In addition to the time of day, StyleDiff found that the two datasets differ in their distributions of the background as shown in Figure 11.
StyleDiff visualizes the differences in attribute distributions as histograms e.g., Figures 10-12 (a).
The differences in attributions are quantified by the Wasserstein distance for each attribute, and developers can know which attribute have the largest differences using the distances.
As shown in Figures 10-12 (c), (d), and (e), the developers can display images at specific regions (e.g., a region with a large difference in the distributions) for each attribute.
This helps the developers to understand what this extracted attribute is.
StyleDiff also generates a sequence of images with only the attribute varied as shown in Figures 10-12 (b).
The image sequence helps the developer to understand what the attribute is about.

\begin{figure}
    \begin{minipage}[t]{\textwidth}
        \centering
        \includegraphics[width=0.99\linewidth]{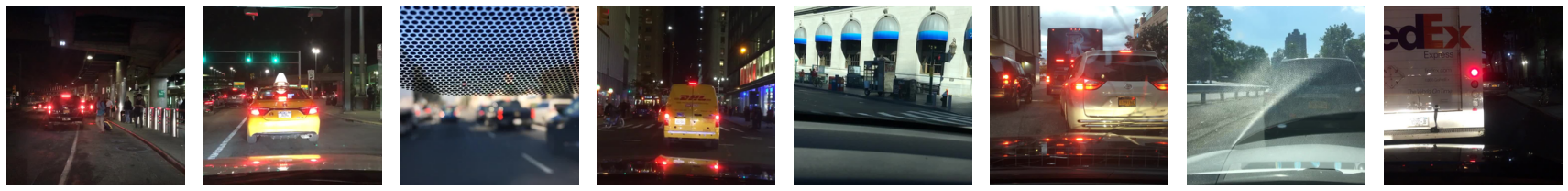}
        \caption{Images selected by PWC~\cite{kawano2022partial}}
        \label{fig:pwc}
    \end{minipage}\\
    \begin{minipage}[t]{\textwidth}
        \centering
        \includegraphics[width=0.99\linewidth]{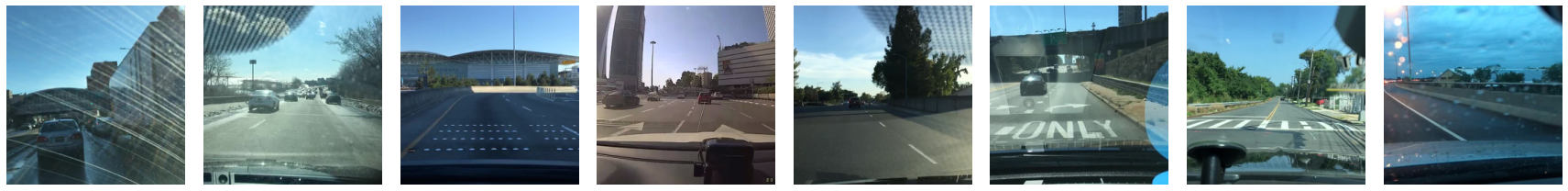}
        \caption{Images selected by LOF~\cite{breunig2000lof}}
        \label{fig:lof}
    \end{minipage}\\
    \begin{minipage}[t]{\textwidth}
        \centering
        \includegraphics[width=0.99\linewidth]{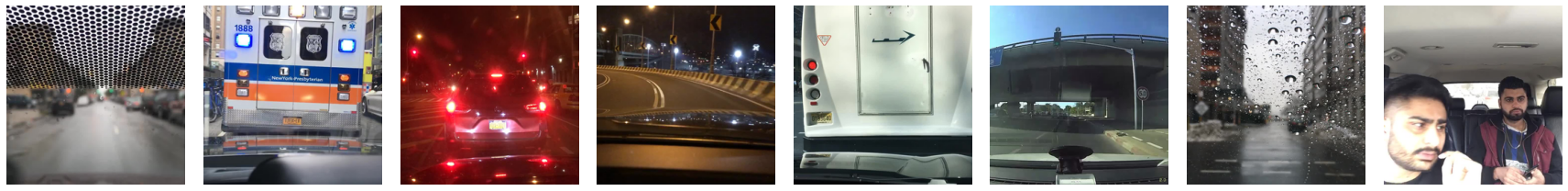}
        \caption{Images selected by coreset (k-center greedy)~\cite{bachem2017practical,sener2018active}}
        \label{fig:coreset}
    \end{minipage}\\
    \begin{minipage}[t]{\textwidth}
        \centering
        \includegraphics[width=0.99\linewidth]{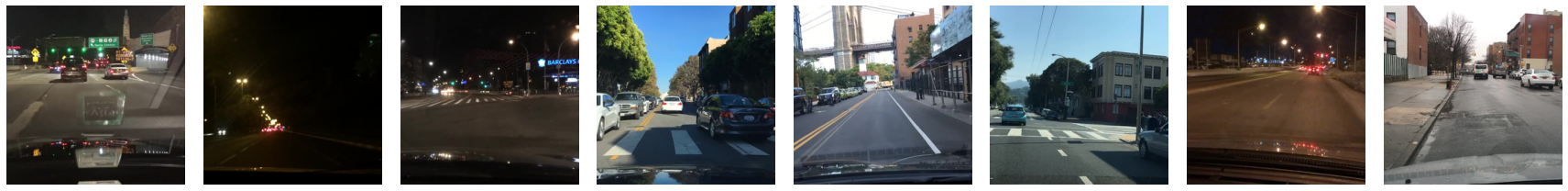}
        \caption{Images selected by FID (FID greedy)~\cite{heusel2017gans, chong2020effectively}}
        \label{fig:fid}
    \end{minipage}
\end{figure}

\paragraph{Results of existing methods}
In Figures~\ref{fig:pwc}, \ref{fig:lof}, \ref{fig:coreset}, and \ref{fig:fid}, we demonstrate the results of existing methods, PWC~\cite{kawano2022partial}, LOF~\cite{breunig2000lof}, coreset~(k-center greedy)~\cite{sener2018active}, and FID greedy~\cite{heusel2017gans, chong2020effectively} on the two datasets, respectively.
Because Figures~\ref{fig:pwc} and \ref{fig:fid} contain several night scenes, users can infer that the KITTI dataset (i.e., the development dataset) has fewer night scenes than the BDD100k dataset (i.e., the real dataset).
However, it is impossible to investigate how the attribute (i.e., daytime and nighttime) is distributed in the two datasets.
Furthermore, it is not even possible to infer if the distributions for the other attributes are different.
From the results of the LOF and coreset, it is difficult to understand the most significant difference (daytime and nighttime) between the two datasets.
We also emphasize that StyleDiff allows us to visualize histograms of attribute distributions, obtain images corresponding to specific regions in the histogram, and generate image sequences corresponding to the attributes, although the information regarding attributes obtained by existing methods is only a set of images.

\begin{figure}
    \centering
    \includegraphics[width=0.99\linewidth]{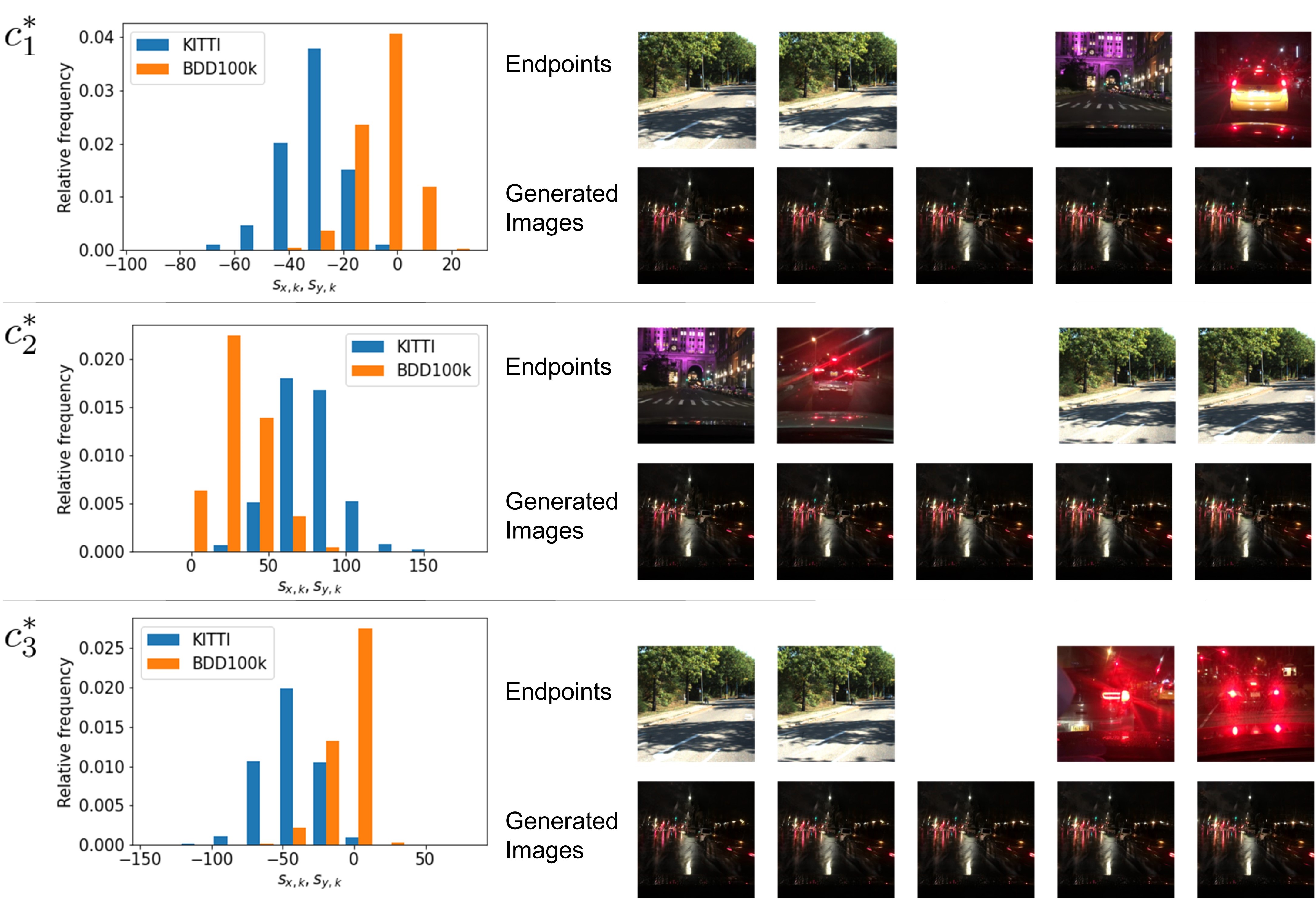}
    \caption{Failure example: comparison results of BDD100k and KITTI if the style vectors are  used directly as the attribute vectors}
    \label{fig:false_results}
\end{figure}

\paragraph{StyleDiff with an attribute space having low completeness}
\label{sec:fail}
Because the performance of StyleDiff depends on the quality of the attribute space, it is difficult to extract attributes if the disentanglement and completeness of the space are low.
The results of StyleDiff with a poor quality attribute space, StyleSpace trained on the driving scene datasets, are presented in Figure~\ref{fig:false_results}, where the dimensions corresponding to the top-3 largest distances are visualized.
As shown in Figure~\ref{fig:false_results}, the endpoints of all the top-3 dimensions correspond to illuminated night scenes and daytime images, indicating that the single attribute corresponds to the multiple dimensions (i.e., low completeness).
Here, the generated images did not change even if the values of the selected dimension were changed so that the corresponding attribute could not be visualized by the generated images.
This is because completeness of the space is low; multiple dimensions must be changed simultaneously to manipulate the attribute in the image, and changing the value of only one dimension cannot significantly affect the generated images.
Note that, on the contrary, if the disentanglement of the attribute space is low, changing the value of a single dimension may change multiple attributes simultaneously, making it difficult to infer the corresponding attributes.

\section{Conclusion}
\label{sec:conclusion}
We proposed StyleDiff to present how two unlabeled image datasets differ in a human-understandable manner.
With the help of the disentangled latent space of the recent generative models, StyleDiff extracts attributes from the datasets where the attributes are not explicitly labeled.
StyleDiff then quantifies the mismatches between the distributions for each extracted attribute, and visualizes attributes with large mismatches.
The computational complexity of StyleDiff is $O(d  N \log N)$, and it can be applied to datasets with over 10,000 images.
Experimentally, StyleDiff outperformed existing methods in detecting the mismatches between datasets by the quantitative evaluation.
Furthermore, we demonstrated that the mismatches between two real-world datasets, (e.g., KITTI and BDD100k), could be discovered and visualized by StyleDiff.
By using StyleDiff, the developers can accurately understand the mismatches between datasets and make appropriate improvements such as data collection, additional testing, and developing a new subsystem.
By these improvements, the performance and reliability of machine learning systems can be steadily improved.
A limitation of StyleDiff is that the performance of StyleDiff depends on the quality of the attribute space because StyleDiff assumes that each dimension of the attribute space has a one-to-one correspondence with an attribute.
If a latent space of a generative model in which each attribute corresponds to multiple dimensions (low completeness) or multiple attributes correspond to a dimension (low disentanglement) is used as the attribute space, StyleDiff will fail to extract attributes individually.
This will cause subsequent processes to fail; quantification of differences between attribute distributions becomes inaccurate and visualization of latent attributes becomes difficult to understand.
A continuation of our work is to use latent spaces obtained from more advanced generative models for the attribute space. 
By using latent spaces with higher completeness and disentanglement, StyleDiff can present mismatches in the datasets more comprehensively and exhaustively.

\appendix
\setcounter{table}{0}

\section{Quantitative Evaluation with CelebA Dataset}
\label{appendix:celeba}
In this section, we performed a quantitative evaluation using the CelebA dataset~\cite{liu2015faceattributes}.
The experimental procedure is the same as that in Section~\ref{sec:quantitative}.
We generated development and real datasets so that the two datasets have different distributions for each of 33 attributes annotated on the images in CelebA dataset.
Using the datasets, we evaluated the performance of StyleDiff and the conventional methods in extracting the attribute with different distributions.
See Section~\ref{sec:quantitative} for the detailed procedures.
Table~\ref{tab:result-celeba} shows the results of the quantitative evaluation.
As shown in Table~\ref{tab:result-celeba}, StyleDiff outperformed the conventional methods on the CelebA dataset~\cite{liu2015faceattributes} as well as on the FFHQ dataset~\cite{karras2019style}.

\begin{table}
    \caption{Quantitative evaluation results with CelebA dataset. SD denotes StyleDiff.}
    \label{tab:result-celeba}
    \centering
    \begin{tabular}{P{3.2cm}P{1.1cm}P{1.9cm}P{1cm}P{1.1cm}P{1.7cm}P{1cm}}
    Attribute Name        & SD             & SD w/o normalize & PWC \cite{kawano2022partial} & LOF \cite{breunig2000lof} & k-center \cite{bachem2017practical,sener2018active} &FID greedy \cite{heusel2017gans, chong2020effectively} \\	     \toprule
5-o-ClockShadow & \textbf{0.865} & 0.845 & 0.806 & 0.724 & 0.756 & 0.789  \\
ArchedEyebrows & \textbf{0.847} & 0.816 & 0.803 & 0.762 & 0.750 & 0.772  \\
BagsUnderEyes & \textbf{0.839} & 0.806 & 0.785 & 0.738 & 0.757 & 0.769  \\
Bangs & \textbf{0.896} & 0.835 & 0.793 & 0.768 & 0.766 & 0.787  \\
BigLips & 0.769 & 0.766 & 0.769 & \textbf{0.776} & 0.748 & 0.752  \\
BigNose & \textbf{0.861} & 0.828 & 0.790 & 0.738 & 0.754 & 0.775  \\
BlackHair & \textbf{0.901} & 0.837 & 0.780 & 0.763 & 0.761 & 0.768  \\
BlondHair & \textbf{0.903} & 0.854 & 0.804 & 0.751 & 0.744 & 0.770  \\
Blurry & \textbf{0.896} & 0.866 & 0.780 & 0.750 & 0.752 & 0.780  \\
BrownHair & \textbf{0.804} & 0.758 & 0.761 & 0.747 & 0.735 & 0.758  \\
BushyEyebrows & \textbf{0.821} & 0.767 & 0.764 & 0.747 & 0.754 & 0.769  \\
Double-Chin & \textbf{0.896} & 0.865 & 0.831 & 0.733 & 0.764 & 0.790  \\
Eyeglasses & \textbf{0.902} & 0.885 & 0.838 & 0.747 & 0.769 & 0.788  \\
Goatee & \textbf{0.916} & 0.896 & 0.819 & 0.712 & 0.748 & 0.781  \\
GrayHair & \textbf{0.919} & 0.875 & 0.834 & 0.734 & 0.761 & 0.782  \\
HeavyMakeup & \textbf{0.900} & 0.884 & 0.838 & 0.769 & 0.758 & 0.787  \\
HighCheekbones & \textbf{0.815} & 0.783 & 0.758 & 0.751 & 0.746 & 0.784  \\
Male & \textbf{0.908} & 0.905 & 0.851 & 0.746 & 0.779 & 0.771  \\
MouthSlightlyOpen & \textbf{0.845} & 0.779 & 0.754 & 0.760 & 0.749 & 0.773  \\
Mustache & \textbf{0.900} & 0.884 & 0.823 & 0.738 & 0.769 & 0.789  \\
NoBeard & \textbf{0.902} & 0.878 & 0.828 & 0.737 & 0.766 & 0.779  \\
OvalFace & \textbf{0.779} & 0.765 & 0.749 & 0.745 & 0.739 & 0.767  \\
PointyNose & 0.792 & \textbf{0.793} & 0.771 & 0.749 & 0.735 & 0.755  \\
RosyCheeks & \textbf{0.871} & 0.845 & 0.761 & 0.735 & 0.702 & 0.780  \\
Sideburns & \textbf{0.902} & 0.875 & 0.826 & 0.739 & 0.768 & 0.796  \\
Smiling & \textbf{0.835} & 0.787 & 0.787 & 0.751 & 0.747 & 0.779  \\
StraightHair & \textbf{0.794} & 0.745 & 0.745 & 0.743 & 0.734 & 0.783  \\
WavyHair & \textbf{0.850} & 0.817 & 0.817 & 0.752 & 0.741 & 0.778  \\
WearingEarrings & \textbf{0.849} & 0.831 & 0.831 & 0.747 & 0.754 & 0.785  \\
WearingHat & \textbf{0.903} & 0.858 & 0.858 & 0.723 & 0.725 & 0.790  \\
WearingLipstick & \textbf{0.917} & 0.900 & 0.900 & 0.742 & 0.745 & 0.775  \\
WearingNecklace & \textbf{0.843} & 0.830 & 0.830 & 0.755 & 0.748 & 0.774  \\
WearingNecktie & \textbf{0.904} & 0.884 & 0.884 & 0.678 & 0.702 & 0.786 
\end{tabular}
\end{table}

\clearpage

\bibliographystyle{elsarticle-num}
\bibliography{references}

\end{document}